\documentclass[10pt,twocolumn,letterpaper]{article}
\usepackage[pagenumbers]{cvpr}
\usepackage{array}\usepackage{colortbl}\usepackage{cuted}\usepackage{enumitem}\usepackage{tikz}
\usetikzlibrary{positioning}
\definecolor{sprited}{HTML}{6337C7}\definecolor{teal}{HTML}{008C95}\definecolor{pale}{HTML}{F3F0FA}\definecolor{muted}{HTML}{555555}
\definecolor{control}{HTML}{F2F2F2}\definecolor{pixel}{HTML}{E9F7F7}\definecolor{latent}{HTML}{F3F0FA}
\definecolor{cvprblue}{rgb}{0.21,0.49,0.74}
\usepackage[breaklinks,colorlinks,allcolors=cvprblue]{hyperref}
\hypersetup{
pdftitle={Dancing Stick Figures: An Introductory Dataset for Training Video Generation Models},
pdfauthor={Jin Hyuk Cho},pdfsubject={Synthetic video dataset, generation harness, and diagnostic evaluation suite},
pdfkeywords={video generation, benchmark, synthetic dataset, structural evaluation, education}}
\setlist{nosep,leftmargin=*}

\def\confName{CVPR}
\def\confYear{2026}

\title{Dancing Stick Figures: An Introductory Dataset\\for Training Video Generation Models}
\author{Jin Hyuk Cho\\Sprited\\
\small\href{mailto:jin@sprited.ai}{jin@sprited.ai}\\[-2pt]
\small\href{https://huggingface.co/datasets/sprited/dancing-stick-figures}{Dataset} $\cdot$
\href{https://github.com/sprited-ai/dancing-stick-figures}{Code} $\cdot$
\href{https://huggingface.co/sprited/dancing-stick-figures-baselines}{Models} $\cdot$
\href{https://colab.research.google.com/github/sprited-ai/dancing-stick-figures/blob/main/notebooks/dancing_stick_figures_colab_v0_3.ipynb}{Colab}}
\date{}

\begin{document}
\maketitle
\begin{strip}
\centering
\includegraphics[width=\textwidth]{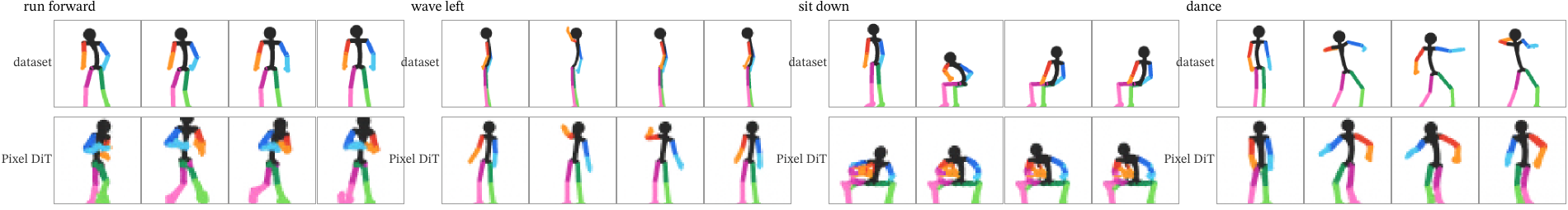}
\captionsetup{hypcap=false}
\captionof{figure}{Released dataset clips and outputs from the 30k-step image-initialized Pixel DiT
for locomotion, gesture, transition, and dance prompts. Each block shows frames 0/21/42/63 from a released
clip and an independently seeded generation (50 Euler steps, guidance 3).}
\captionsetup{hypcap=true}
\label{fig:dit-samples}
\end{strip}
\begin{abstract}
Training a video-generation model from scratch is hard for reasons that precede model design. The feedback
loop is long: a failure that appears only after a training run can make each attempted fix another run. The
data are hard to reach: the corpora and recipes behind strong models are large, heterogeneous, and often
unreleased. And scoring is blunt: open-ended generation has no single correct output, and an aggregate score
does not by itself establish whether a sample succeeds or which property failed. \textbf{Dancing Stick
Figures} is a synthetic video dataset built against these three obstacles. For iteration speed, its $64^2$,
64-frame reference task is sized for practical repeated training on a single workstation GPU. For accessibility, the
release is a 0.79-GB training tier of 4,020 video clips---1,340 six-second source motions, each rendered from
three cameras by a deterministic dataset-generation harness---with checkpoints and a Colab workflow that
reruns the reference training pipeline at reduced budget on a 16\,GB Tesla T4. For scoring, every frame
retains its generating state (ARDY cskel27 joint positions, camera, body parameters, and source motion) and
per-pixel depth, surface normals, and part labels. These annotations support dataset-specific metrics for
visible topology and part-wise motion; corruptions expose their sensitivities and blind spots.
\end{abstract}

\section{Introduction}
Training a video-generation model from scratch takes a heavy toll on the wallet. The reference recipes
assume GPU servers, and renting them turns every experiment into a billing decision---a poor mode of study.
Study wants a small, annotation-rich dataset one can play with: download it, inspect it, train on it
repeatedly, and change one thing at a time. Public datasets of real footage exist at every scale, but the
ones rich enough to train a video model are typically too large for repeated from-scratch training, and
collected footage records nothing about the true scene behind each frame. What is missing is a dataset
purpose-built for research: small enough to fit one GPU, fast enough to see an end-to-end result within a
day, and procedural, so that a researcher can tweak the generator and regenerate the data---and so the
generating parameters themselves can seed score functions. We built Dancing Stick Figures for that role.
Three barriers shape its design:
\begin{itemize}[leftmargin=*,nosep]
\item \textbf{Iteration cost.} The complete training path for frontier systems such as Seedance and MiniMax
H3~\cite{gao2025seedance,minimax2026h3} runs through large corpora, heavy compute, pretrained codecs, and often
proprietary artifacts; controlled retraining there is impractical. A failure that appears only after
training---for example, motion that becomes static late in a clip---can make each attempted diagnosis
another expensive run.
\item \textbf{Accessibility.} Public datasets, training code, pretrained components, and benchmarks are
available, but rarely as one documented, runnable workflow. A researcher must still choose compatible
components, reconstruct preprocessing and training details, and determine what results to expect. This
integration burden makes it difficult to begin with a small experiment and modify it with confidence.
\item \textbf{Diagnosis.} Open-ended generation has no single correct output, while failures differ: motion
may become static, flicker, or lose smoothness; subject appearance may drift; and the action may miss the
prompt. These correspond to distinct evaluation dimensions such as dynamic degree, temporal flickering,
motion smoothness, subject consistency, and prompt adherence~\cite{huang2023vbench}; one aggregate score
cannot identify which property failed.
\end{itemize}

These barriers lead to three corresponding design choices: a compact reference task for repeated training, a
connected release that runs end to end, and procedural data whose generating state supports
dataset-specific diagnosis.

The procedural design exposes variables that are usually unavailable in collected video. Each rendered frame
is linked to its source motion, rig and body, camera, renderer, depth, surface normals, and part labels. We
can change one variable at a time, rerender the clip, and observe which diagnostics react. Claims from
these interventions are limited to the recorded variables and controlled changes tested here.

Can a researcher train reference models from scratch on a single workstation GPU and then iterate on them through
short, controlled cycles? Each cycle should expose a named failure, change one mechanism, and rerun the
same diagnostic. The release supports that workflow with a connected end-to-end route, an image checkpoint
used to initialize the video model's spatial weights, runs short enough to repeat and vary, and diagnostics
with known blind spots.

Dancing Stick Figures connects three operations: regenerate the dataset from its recorded source parameters,
train a 64-frame prompt-conditioned video generator, and evaluate its outputs with validated dataset-specific
metrics.

The release contributes:
\begin{enumerate}
\item a label-rich video dataset released at $128^2$ and in a 0.79-GB $64^2$ tier, with the motion, camera,
and body parameters behind every frame, and a deterministic dataset-generation harness;
\item a seed-disjoint evaluation protocol and dataset-specific topology and motion metrics, validated with a
controlled corruption suite; and
\item image and video trainers, checkpoints, and a reduced-budget Colab workflow that reruns the reference
training pipeline end to end.
\end{enumerate}

\begin{figure*}[t]\centering
\begin{tikzpicture}[font=\small,node distance=6mm and 6mm,
art/.style={draw=sprited,rounded corners=2pt,fill=pale,align=center,minimum height=8mm,inner xsep=6pt},
op/.style={draw=teal,thick,rounded corners=2pt,fill=white,align=center,minimum height=8mm,inner xsep=6pt},
diag/.style={draw=teal,rounded corners=2pt,fill=teal!10,align=center,minimum height=8mm,inner xsep=6pt},
corr/.style={draw=muted,dashed,rounded corners=2pt,fill=white,align=center,minimum height=8mm,inner xsep=6pt},
grp/.style={font=\scriptsize\itshape,text=muted},
arr/.style={-latex,thick,draw=muted},
darr/.style={-latex,thick,draw=muted,dashed}]
\node[art] (src) {134 prompts\\1,340 source motions};
\node[op,right=of src] (harness) {dataset-generation harness\\body + camera + rendering};
\node[art,right=of harness] (data) {4,020 video clips\\state + depth/normals/parts};
\node[op,below=9mm of harness] (train) {reference training\\image pretraining\\64-frame video training};
\node[art,right=of train] (ckpt) {checkpoints\\+ generated samples};
\node[diag,right=of ckpt] (eval) {evaluation signals\\topology + part-wise motion\\secondary: FVD};
\node[corr,below=6mm of eval] (corrupt) {controlled corruptions\\rig edits + frame order};
\draw[arr] (src)--(harness);
\draw[arr] (harness)--(data);
\draw[arr] (data.south) -- ++(0,-3.5mm) -| (train.north);
\draw[arr] (train)--(ckpt);
\draw[arr] (ckpt)--(eval);
\draw[arr] (data.east) -| (eval.north) node[grp,pos=0.5,anchor=north west,xshift=1.5mm,align=left]{recorded state\\+ real references};
\draw[darr] (corrupt.north)--(eval.south) node[grp,midway,right=1mm]{stress-test sensitivity};
\node[grp,above=1mm of src.north west,anchor=south west] {data construction};
\node[grp,below=1mm of train.south west,anchor=north west] {reference experiment};
\node[grp,below=1mm of corrupt.south west,anchor=north west] {diagnostic validation};
\end{tikzpicture}
\caption{System map of the release. The released dataset is an artifact of the dataset-generation harness
applied to the released prompts and source motions. Reference training consumes the dataset and produces
model artifacts: checkpoints and generated samples. Recorded state and real references from the dataset,
together with generated samples, feed the evaluation signals; controlled corruptions validate what those
signals detect and miss.}\label{fig:loop}
\end{figure*}
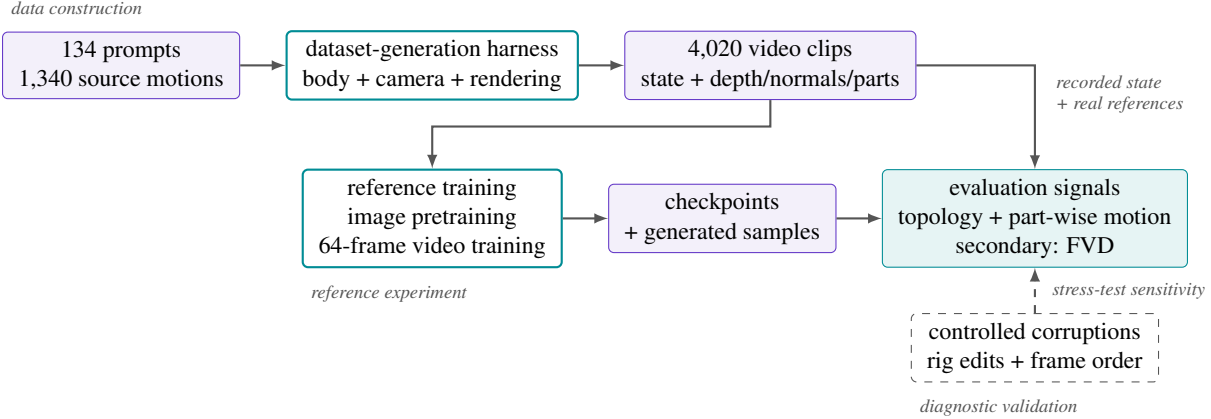

\section{Related Work}

\paragraph{Controlled synthetic data.} Moving MNIST~\cite{srivastava2015unsupervised} made temporal
prediction cheap to study, while Kubric and MOVi~\cite{greff2022kubric} programmatically render scenes with
dense ground truth for scene understanding. SURREAL~\cite{varol2017surreal} and BEDLAM~\cite{black2023bedlam}
extend synthetic rendering to articulated people at a scale and realism aimed at perception. Dancing Stick
Figures focuses on text-conditioned videos of a single stick figure. The small, controlled domain makes
end-to-end retraining affordable and lets us test the failure diagnostics with known corruptions.

\paragraph{Video training data and motion sources.} HumanVid~\cite{wang2024humanvid} and
MiraData~\cite{ju2024miradata} broaden the coverage of video training data, while
AMASS~\cite{mahmood2019amass} and HumanML3D~\cite{guo2022humanml3d} release motion representations. Dancing
Stick Figures connects these layers in one compact release: text-conditioned source trajectories are
rendered into video while preserving the correspondence among prompts, joints, camera and body parameters,
limb labels, and pixels. ARDY~\cite{zhao2026ardy} supplies the upstream 27-joint trajectories; it is not
evaluated by our video diagnostics. This pairing lets the released pixels train a video model, the known
limb structure define dataset-specific measurements, and controlled corruptions test what those measurements
detect.

\paragraph{Evaluation of generated video.} FVD~\cite{unterthiner2018fvd} is widely reported as a
distributional comparison in video-generation benchmarks~\cite{ho2022vdm}. It is not a
diagnostic of whether an individual video succeeds: its I3D features have a documented content bias and can
be under-sensitive to temporal corruption~\cite{ge2024fvdbias}. Our controlled study uses FVD as a
conventional coarse reference and tests that limitation directly in this domain.
VBench~\cite{huang2023vbench} demonstrates the value of separating video quality into diagnostic dimensions.
Our setting adds recorded generating state, which lets us apply known corruptions and test which
dataset-specific signals respond.

\section{Dataset and generation pipeline}\label{sec:data}
We generated one motion per seed 0--9 for each of 143 prompts in six groups (dance, gesture, locomotion,
transition, idle, sport) using NVIDIA ARDY~\cite{zhao2026ardy}. Its Core model returns batchable 20-fps
trajectories on a fixed 27-joint skeleton, matching the harness input without skeleton conversion. Each
trajectory retains its prompt association, while body shape, camera, and raster appearance are applied
later under deterministic control. Each motion is six seconds at 20 fps.

ARDY supplies only the source motion; the reference models in this paper generate rendered video and are
evaluated with the video diagnostics. The released motion table lets users rebuild or vary the rendered
dataset without running ARDY. For additional source-motion generation, we use ARDY v0.2.0. The released motion table remains the canonical input for
reproducing the published dataset. The resulting motion distribution and prompt response inherit ARDY's
capabilities and biases. Licensing and provenance are detailed in \S\ref{sec:access}.

A capsule renderer with $4\times$ supersampling emits three deterministic orthographic views.
The release records 143 candidate prompts; nine are marked for exclusion with per-prompt reasons, leaving
134 in the dataset. Selection is at prompt level, so all seeds of retained prompts remain. Table~\ref{tab:data} summarises the release. Seeds 0--7 train, seed
8 validates, and seed 9 tests: every released prompt appears in all three splits, and no source motion
crosses a split boundary (Table~\ref{tab:splits}). Because selection considered realisations across seeds,
held-out evaluation measures unseen motion realisations of known, selected prompts, not unseen-prompt
generalisation.

\begin{table}[t]\centering\caption{Dataset configurations and storage footprint. A motion is rendered from three views; \texttt{frames} and
\texttt{mini} store one record per rendered frame, \texttt{motion} one record per source
motion.}\label{tab:data}\small
\begin{tabular}{@{}lrrr@{}}\toprule Configuration & Records & Resolution & Size \\\midrule
\texttt{frames} & 482,400 & $128^2$ & 4.4 GB \\
\texttt{mini} & 482,400 & $64^2$ & 0.79 GB \\
\texttt{motion} & 1,340 & 27 joints & 0.33 GB \\\bottomrule\end{tabular}\end{table}

\begin{table}[t]\centering
\caption{Seed-disjoint splits. Each split contains all 134 prompts; source motions are disjoint by seed.}
\label{tab:splits}\scriptsize
\begin{tabular}{@{}lrrrr@{}}\toprule
Split & Prompts & Motions & Videos & Frames \\\midrule
Train & 134 & 1,072 & 3,216 & 385,920 \\
Validation & 134 & 134 & 402 & 48,240 \\
Test & 134 & 134 & 402 & 48,240 \\
All & 134 & 1,340 & 4,020 & 482,400 \\\bottomrule
\end{tabular}
\end{table}

\begin{figure*}[t]
\centering
\includegraphics[width=\textwidth]{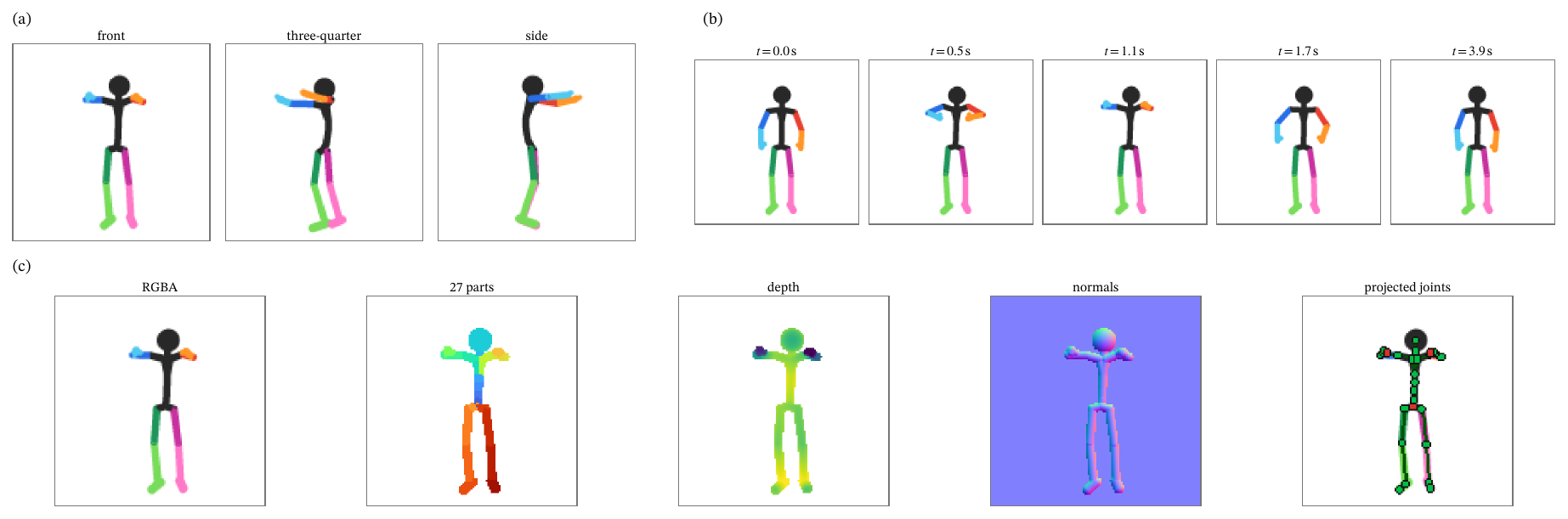}
\caption{One released motion---\texttt{A person waves goodbye with both hands.}---viewed through the
dataset. (a) The same representative pose from front, three-quarter, and side cameras. (b) Five frames at
equal increments of pose-space motion, so an idle tail cannot dominate the strip. (c) The same frame with
part segmentation, depth, camera-space normals, and projected joints. Each record also stores its prompt,
source-motion identifier, seed, camera, and body parameters.}
\label{fig:dataset-anatomy}
\end{figure*}

The generator samples limb length, scale, stroke, and camera yaw and pitch deterministically from
\texttt{clip\_id}. It stores 3D and projected joints, visibility, camera and body parameters, RGBA, 16-bit
depth, camera-space normals, 27-part segmentation, and the raw motion (Figure~\ref{fig:dataset-anatomy});
the dataset card is the schema reference. Released rendering parameters span 50--58 pixels/m, 3--5-pixel
strokes, $-3^\circ$--$10^\circ$ camera pitch, and .92--1.08 per-bone scale multipliers. The six groups
contain 33 dance, 29 gesture, 21 locomotion, 13 transition, 10 idle, and 28 sport prompts. The builder
records two motion-quality flags: \texttt{frozen} (mean joint speed below 0.02\,m/s) and \texttt{levitation}
(root ever above 1.6\,m). Of 1,340 source motions, 58 are marked frozen (41 from the idle group) and one
levitation; flagged samples stay in the release so filtering choices are explicit. Across 4,020 rendered
clips, the median foreground fraction is .059 (5th--95th percentiles .043--.077). After deduplicating views
by retaining camera 0 once per source motion, median centroid speed is .269 pixels/frame (.058--.635) and
summed projected-rig part angular path is 24.75 rad (2.37--94.99). The largest train--validation or
train--test Kolmogorov--Smirnov distance across these three quantities is .106.

The released configurations are two resolutions of one dataset: $128^2$ keeps the most detail, and the 0.79-GB
$64^2$ tier is the primary training tier; a derived $32^2$ cache view supports smoke tests. Reported structural
results use $64^2$, where the thin colored limbs remain measurable. The motion release supports rerendering
the released motions into the full visual dataset without ARDY. \S\ref{sec:access} describes the rebuild
and its verification.

\section{Reference benchmark and evaluation}
\paragraph{Generation setup.} We use the dataset to define a compact reference benchmark for
prompt-conditioned video generation. Each public clip has 120 frames at 20 fps. Every benchmark and
diagnostic experiment in this paper uses the fixed first 64-frame window at native 20 fps, matching the
reference-model output. The reference models train on and generate the fixed first 64 frames of each clip
(3.2 seconds). The window rows in
Table~\ref{tab:backbone-results} compare generated 64-frame windows with real windows drawn under the same
manifest rule.

\begin{figure*}[t]\centering
\includegraphics[width=0.98\textwidth]{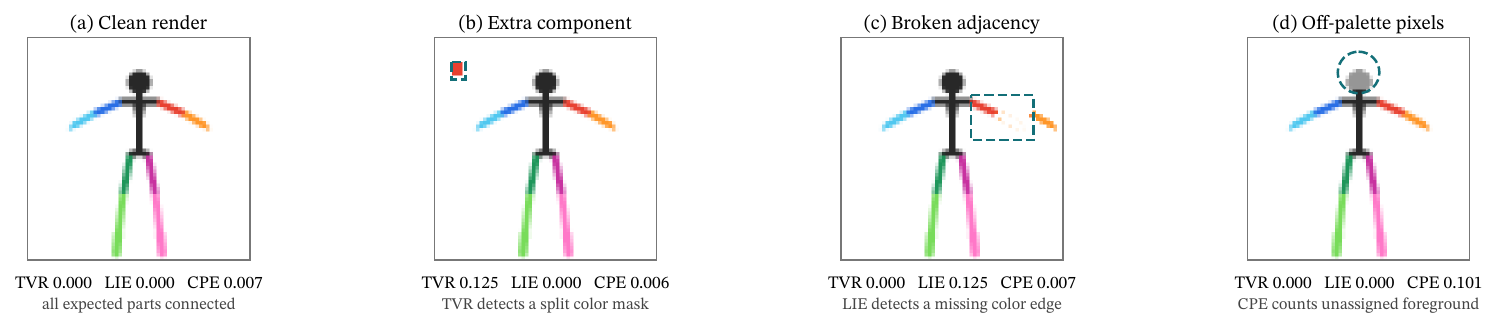}
\caption{Controlled examples of the three structural scores at the released $64^2$ evaluation resolution. All
four panels derive from the same rendered frame and are scored by the released evaluator. A detached
limb-color component raises TVR by $1/8$; translating one distal color mask beyond the two-pixel adjacency
tolerance raises LIE by $1/8$; recoloring foreground pixels outside the palette raises CPE. The clean CPE is
non-zero because some antialiased boundary pixels fall outside the RGB-distance tolerance. These edits
illustrate the score definitions; they are not generated-video samples.}
\label{fig:topology-metric-examples}
\end{figure*}

\paragraph{Structural diagnostics.} Each arm and leg uses a fixed pair of colors. The evaluator converts a
frame into eight limb-color masks and reports:
\begin{itemize}\setlength\itemsep{0pt}
    \item \textbf{Topology violation rate (TVR):} limb colors missing or split into multiple components;
    \item \textbf{Limb-identity error (LIE):} expected color adjacencies that are absent;
    \item \textbf{Color-purity error (CPE):} foreground pixels outside the rendering palette;
    \item \textbf{Foreground fraction:} pixels occupied by the figure; and
    \item \textbf{Clean rate:} frames with both TVR and LIE at zero.
\end{itemize}
Implementation uses 8-connected components (a 3-by-3 neighborhood) for TVR. For LIE, an expected color
pair is adjacent when two iterations of 3-by-3 dilation of the proximal mask overlap the distal mask, giving
a two-pixel Chebyshev tolerance. Foreground means alpha above 127; palette assignment requires Euclidean RGB
distance below 60, and a color mask with fewer than four assigned pixels is treated as absent. Video-level
TVR, LIE, and CPE are means over frames.
On one deterministic $64^2$ frame from each of the 4,020 released clips, threshold 60 assigns 96.4\% of the
evaluator's alpha-gated foreground pixels and reaches 98.2\% class accuracy among assigned pixels and
91.1\% macro IoU against the released segmentation mapped to the evaluator classes. In a threshold sweep
from 20 to 100, threshold 60 is within 0.1 percentage point of the best macro IoU; the alpha${>}127$
evaluator foreground gate excludes 17.2\% of segmentation-labeled foreground pixels, a
foreground-definition scope difference rather than a classification error.
Figure~\ref{fig:topology-metric-examples} shows one isolated trigger for each score.
These are dataset-specific metrics that exploit the fixed limb colors: they query specific structural
properties of the rendered figure, and none is a complete correctness judgement. For video, the evaluator adds
color-mass drift, centroid speed and acceleration, motion fraction, angular speed and jerk, and height
variation; some track the whole foreground, others individual limbs. Centroid speed is the mean foreground-centroid displacement in
pixels per frame; motion fraction is the fraction of frame transitions whose centroid displacement exceeds the
released 0.25-pixel threshold; angular jerk averages the absolute second differences of each limb's
principal-axis angle. These are image-space diagnostics, not biomechanics. The motion quantities are
two-sided reference signals: zero can mean a stopped video, and large values can mean jitter. Agreement with
the real reference, not a small value, is the target.

Occlusion produces non-zero topology scores even on real rendered frames, so scores on real rendered frames
are a \emph{real reference}, not an error floor. A deterministic manifest selects one camera per source
motion---so correlated views of the same motion never enter as independent samples---and keeps the two
128-video reference halves motion-disjoint. For completeness, we report FVD~\cite{unterthiner2018fvd} as a
secondary coarse distributional reference, not a primary measure of video quality or a diagnostic score. We
composite RGBA onto white, resize to $224^2$, and extract 400-dimensional features with the standard
Kinetics-400 I3D TorchScript network. Because that network was trained on natural action videos and this
study has only 128 held-out source motions, values are comparable only within this fixed implementation,
window length, and sample count; close differences require uncertainty analysis.

\paragraph{Controlled validation.} Five deterministic structural corruptions were applied to 500 held-out
frames re-rendered from their stored labels, and scored against the same frames uncorrupted. A partial
left/right limb swap raises LIE from .063 to .266; an extra arm raises TVR from .176 to .261. A whole-limb
swap preserves the color graph and is not detected; bone stretching and hand deletion are also missed, because
the metrics measure topology, not proportions or end-effectors. Temporal corruptions (freeze, shuffle, loop,
reverse) are evaluated in \S\ref{sec:results}.

\begin{figure*}[t]\centering
\includegraphics[width=0.96\textwidth]{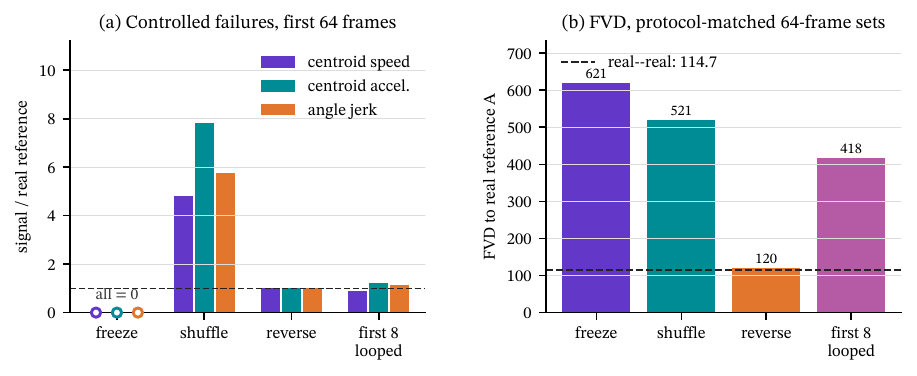}
\caption{Controlled study on protocol-matched first-64-frame windows ($n=128$ per set). \emph{Freeze}
repeats window frame 1, \emph{shuffle} permutes the 64 selected frames, \emph{reverse} reverses the order of
those same 64 frames, \emph{first 8 looped} repeats window frames 1--8. (a) Freezing zeroes the motion
signals; shuffling and looping create excess acceleration and angular jerk; reversal changes neither.
(b) Against the 114.7 real--real reference, FVD scores freezing 620.8, shuffling 520.7, reversal 120.1, and
looping 418.2.}
\label{fig:failures}\end{figure*}

\section{Diagnostic validation and reference results}\label{sec:results}

\paragraph{Metric stress test.} Figure~\ref{fig:failures} reports the 64-frame study on protocol-matched
first-64-frame windows ($n=128$ per set). The real reference half B has centroid speed .373, centroid
acceleration .415, and angular jerk .073. Freezing zeroes all motion signals. Shuffling raises the three
signals to 1.802, 3.244, and .420; looping the first eight frames gives .336, .498, and .083. Reversal
changes neither the frame set nor any of the time-symmetric hand-designed signals. FVD scores the
corruptions against the 114.7 real--real reference at 620.8 for freezing, 520.7 for shuffling, 418.2 for
looping, and 120.1 for reversal. Across 30 paired resamples of 64 videos from the 128-video source set, the
paired FVD difference under reversal is $5.24\pm5.24$ (mean $\pm$ standard deviation; range $-5.76$ to
$15.74$), positive in 25/30 resamples. The hand-designed signals are reversal-invariant by construction,
and this FVD responds only weakly and inconsistently to reversal; neither certifies directionally correct
motion.

\paragraph{Part-wise localization check.} We applied a topology-preserving arm-chain freeze at four
severities to 24 unflagged test motions, using camera 0 once per source motion. For each clip, the arm with
greater clean rendered angular path was selected; its local bone directions were blended toward their
first-frame values while attachment and bone lengths were preserved. The median reduction in the two
affected color-part angular paths increased from 1.51 to 4.69, 8.61, and 9.11 rad across severities .25,
.5, .75, and 1.0. Median absolute change across the other six limb parts remained .006--.011 rad; the
affected reduction exceeded that change in 18/24, 23/24, 22/24, and 23/24 clips. At full severity, the
95\% clip-bootstrap intervals are 4.58--16.43 rad for affected reduction and .001--.032 rad for unaffected
absolute change. Median absolute TVR and LIE changes at full severity are .031 and .001, respectively: the
part-wise signal localizes this arm-chain edit, while severe pose changes can still alter visible topology
through occlusion.

\paragraph{I3D embedding sanity check.} We validated the released FVD protocol at the distribution level on
the final v0.2 release (camera 0 only, first 64 frames at the native frame rate, 134 prompts $\times$ 10
seeds), using ten deterministic matched-$N$ partitions with $N{=}650$ videos per set. Splitting the same
prompt roster by motion seed gives median FVD 26.37 (range 23.72--30.13), close to the random-video-half
baseline of 24.61 (21.39--28.03), while disjoint group-balanced prompt rosters give 47.82 (39.51--57.32);
the same-roster value was lower in 10/10 partitions. FVD therefore responds to dataset-level
action-composition change and is relatively insensitive to seed variation, but it is not a per-video or
prompt-correctness score. We also tested the per-video I3D features directly, as pairwise Euclidean
distances rather than FVD. Among 256 clean clips, 123 same-prompt, different-seed pairs have median
distance 24.21 against 32.82 for matched different-prompt pairs within the same broad motion group; a
same-prompt pair is closer in 80.4\% of balanced comparisons, and nearest-neighbor prompt identity is
18.4\% against a .38\% random baseline. Paired clean-to-corruption median distances are 29.09 for freezing,
27.52 for shuffling, and 28.46 for looping, but only 9.34 for reversal: the embedding carries prompt-linked
action information and reacts to several temporal corruptions, while reversal remains comparatively weak.

\paragraph{Training objective.} Every released denoiser uses velocity-prediction flow matching on
$x_t=(1-t)x_0+t\epsilon$, with $t=\mathrm{sigmoid}(u)$, $u\sim\mathcal N(0,1)$, and target
$v^*=\epsilon-x_0$ (often called rectified flow, though we apply no reflow step)~\cite{peebles2023dit,lipman2023flow}.
For the Image DiT, the loss is
$\mathcal L_{\mathrm{pixel}}=\mathrm E[(1+\alpha)\|\hat v-v^*\|_2^2]/\mathrm E[1+\alpha]$,
where the ground-truth soft alpha is shared across the four RGBA channels. The three pixel-video DiTs in
Table~\ref{tab:backbone-results} add an RGBA clean-prediction auxiliary of weight 1. From
$\hat x_0=x_t-t\hat v$, it penalizes alpha disagreement, foreground RGB squared error, and a 0.1-weighted
background RGB error. The term is gated by $(1-t)$ and uses
$w=1+9\,\mathrm{clip}(a+|a-\hat a|,0,1)$ to emphasize visible and mispredicted foreground. Mini-Wan applies
the corresponding observable-space terms after its frozen codec; because its primary loss is latent-space
flow matching, its objective is analogous but not mathematically identical. Each model uses a frozen
T5-small encoder attached through cross-attention. These rows are reference points for users of the dataset.
\begin{itemize}[leftmargin=*,nosep]
\item \textbf{Image DiT.} The 39.9M-parameter factorized backbone ($4\times4$ frame patches, alternating
spatial and temporal attention blocks) run with $T{=}1$; its checkpoint supplies the spatial and text
weights for image initialization, an established practice~\cite{ho2022vdm,singer2023makeavideo}.
\item \textbf{Pixel video DiTs.} The same 39.9M factorized backbone with $T{=}64$. The video rows separate
random versus image-checkpoint initialization and, among image-initialized models, factorized attention versus a
local $3{\times}3{\times}3$ convolutional mixer (a zero-initialized depthwise-plus-pointwise 3D convolution
per block, about 1.9M extra parameters).
\item \textbf{Dancing Stick Figures Mini-Wan 40M.} Our strongest released video reference has 39.4M
trainable denoiser parameters and uses the Diffusers \texttt{WanTransformer3DModel}, retaining its Wan block
stack: full self-attention over latent video tokens with 3D rotary position encoding and text
cross-attention~\cite{wan2025}. The scale and interface
hyperparameters are adapted: width 384, 16 layers, 6 heads, $(1,2,2)$ patches, and 8 latent channels on a
$16\times16\times16$ grid, with frozen T5-small conditioning. It is trained from random weights and uses no
pretrained Wan weights. For its latent velocity MSE, an alpha-derived occupancy $a_\ell\in[0,1]$ gives each
latent cell weight $1+a_\ell$, normalized by the mean weight. An auxiliary term of weight 1 decodes both
the predicted clean latent and the target latent through a separately trained frozen video codec. Let
$a$ and $\hat a$ be their decoded alpha values and $w=1+9\,\mathrm{clip}(a+|a-\hat a|,0,1)$. The term,
gated by $(1-t)$, adds mean-normalized $w|a-\hat a|$, squared RGB disagreement weighted by
$wa\hat a$, and a 0.1-weighted background RGB penalty. The codec uses fourfold temporal and spatial
compression with eight latent channels (f4t4d8). This is a compact Wan-architecture reference, not a
reproduction of Wan2.1 at its published scale, codec, text encoder, data, or training recipe. We release the
30k checkpoint as the primary latent-space reference.
\item \textbf{Codec control.} Real windows encoded and decoded by that same f4t4d8 codec, separating codec
cost from model skill.
\end{itemize}

\paragraph{Optimization and conditioning mixture.} All learned rows use AdamW, a $2\times10^{-4}$ peak
learning rate after 1{,}000 warm-up updates, cosine decay, and 10\% text-conditioning dropout. The image
stage uses batches of 128; the video stages use effective batches of 16. Each learned video row runs for
30{,}000 optimizer updates, or 480{,}000 sampled examples and 37.3 passes through the training loader.
Pixel-video training draws a single-frame batch with probability .1; the remaining batches are
text-to-video, with no first-frame input. Mini-Wan likewise draws a single latent-time batch with
probability .1 and has no image-conditioning input.

An audit of the frozen f4t4d8 codec bounds its role: held-out foreground reconstruction MSE is 9.5\% higher
than training-prompt MSE, and neutralising background latent cells raises foreground error $5.5\times$. The
mini-Wan row therefore includes both generator and codec error, and it cannot support a general
pixel-versus-latent claim.

Table~\ref{tab:backbone-results} collects the runs. Each block names its protocol, because single-frame and
64-frame scores are different measurements. Earlier v0.1 baselines used a superseded split and are not
comparable rows. Figure~\ref{fig:dit-samples} shows the 30k-step image-initialized factorized Pixel DiT.

\begin{table*}[t]
\centering
\caption{Reference comparisons on the released seed-disjoint data. The three Pixel DiTs use the same
training data, 30k optimizer updates, effective batch 16 (480k sampled examples; 37.3 loader epochs),
optimizer and conditioning mixture, and alpha-weighted pixel-flow objective plus the RGBA auxiliary in the
text. ``Image'' initialization copies compatible spatial and text weights from the single-frame checkpoint.
Mini-Wan uses the same exposure but a different representation, backbone, and exact auxiliary construction,
so it is separated as a latent-space reference. The codec control involves no training. Params count
trainable denoiser parameters and exclude the frozen text encoder and, for Mini-Wan, the separately trained
codec.}
\label{tab:backbone-results}
\scriptsize
\setlength{\tabcolsep}{2.8pt}
\begin{tabular}{@{}lcllrrrrrrr@{}}\toprule
Model & Params & Attention & Init. & TVR$\downarrow$ & LIE$\downarrow$ & CPE$\downarrow$ & Speed & Motion frac. & Jerk & FVD$\downarrow$ \\\midrule
\rowcolor{control}\multicolumn{11}{@{}l}{\textit{Single-frame initializer: prompt-conditioned, $n=512$, 50 sampling steps}}\\
Real validation frames & --- & --- & --- & .113 & .092 & .039 & --- & --- & --- & --- \\
Image DiT, 30k & 39.9M & factorized & random & .128 & .051 & .030 & --- & --- & --- & --- \\
\addlinespace
\rowcolor{control}\multicolumn{11}{@{}l}{\textit{64-frame controls: 128 seed-disjoint source animations; learned rows use three sampling seeds}}\\
Real reference windows & --- & --- & --- & .116 & .093 & .037 & .373 & .501 & .073 & 114.7 / ref. \\
Codec reconstruction floor & --- & --- & --- & .153 & .118 & .053 & .383 & .507 & .106 & 127.1 \\
\addlinespace
\rowcolor{pixel}\multicolumn{11}{@{}l}{\textit{Loss-matched Pixel DiTs: 30k updates, 480k examples, 37.3 loader epochs}}\\
Video DiT, 30k & 39.9M & factorized & random & .259 & .042 & .032 & .366 & .431 & .172 & 483.7 \\
Video DiT, 30k & 39.9M & factorized & image & .161 & .033 & .028 & .350 & .447 & .151 & 319.0 \\
Video DiT + local mixer, 30k & 41.8M & factorized + local mixer & image & .149 & .050 & .035 & .355 & .470 & .146 & 282.9 \\
\addlinespace
\rowcolor{latent}\multicolumn{11}{@{}l}{\textit{Latent reference: 30k updates, 480k examples, 37.3 loader epochs}}\\
Mini-Wan 40M, 30k & 39.4M & Wan full 3D & random & .199 & .101 & .058 & .423 & .534 & .148 & 182.6 \\\bottomrule
\end{tabular}
\par\vspace{2pt}\raggedright\scriptsize TVR is topology violation rate, LIE limb-identity error, CPE
color-purity error. Speed, motion fraction, and jerk are two-sided signals: agreement with the real row is the
target. The codec-floor row passes the real windows through the frozen f4t4d8 codec used by the Mini-Wan
rows; its score should be read against them. FVD is included only as a secondary coarse distributional reference.
The real control row is reference half B; learned-row FVD compares generated videos with reference half A,
and the codec control reconstructs half B and compares it with half A. The two halves are source-motion-disjoint.
The 128 source animations are the distinct reference units, not the total number of generated videos. Each
learned video row contains 384 outputs---one generation for each reference prompt under each of three sampling
seeds. The structural and temporal scores pool those outputs; FVD is computed separately for each 128-video
seed set and then averaged.
With $n=128$ and I3D features trained on Kinetics-400, it is not a calibrated stick-figure quality score and
should not by itself rank nearby rows. Real training-split windows score FVD 125.5 against the same
reference, which is also what a training-set replay would score, so models near this band need a memorisation
check. All table and diagnostic FVD values directly composite premultiplied RGBA over white before I3D
embedding.
\end{table*}

\paragraph{Reading the comparisons.} At matched exposure, image initialization improves the factorized
Pixel DiT from TVR .259 to .161 and FVD 483.7 to 319.0. The three Pixel rows have speed .350--.366, close to
the real-reference .373, while their motion fractions (.431--.470) remain below the real .501. The local
mixer gives the lowest Pixel TVR (.149), jerk (.146), and FVD (282.9); the plain image-initialized
factorized model gives lower LIE (.033) and CPE (.028), so neither dominates every diagnostic. Mini-Wan
reaches FVD 182.6 but changes representation, backbone, and exact auxiliary construction, and its LIE and
CPE are higher. Each learned row is one training run, so these are observed outcomes rather than estimates
of architecture effects. Figure~\ref{fig:dit-samples} shows the corresponding 30k-step image-initialized checkpoint
across locomotion, gesture, transition, and dance prompts.

\paragraph{Why pixel space.} For this release, at $32^2$--$64^2$, pixel-space training is practical and keeps
thin-limb errors in the representation the evaluator scores. A separately trained video VAE adds a training
stage and, as the audit of this released codec shows, can confound generator errors with codec artifacts. The
core Colab therefore follows the pixel-space route.

\section{Accessibility and reproducibility}\label{sec:access}
The repository releases the dataset-generation harness, cache builder, trainers, sampling code, evaluator, corruptions, FVD
evaluation, checkpoints, and comparison script. The Colab notebook is a reduced-budget reproduction of the
reference two-stage training pipeline, not a reproduction of the released checkpoints' quality: with the same code
and architecture it builds a cache, trains the image generator for 2k steps, transfers its spatial weights,
trains the video generator for 2k steps, samples a multi-second GIF, and compares scores against real
references. The Table~\ref{tab:backbone-results} reference checkpoints use 30k updates per stage. $32^2$ is
a derived smoke-test view; $64^2$ is the reference dataset tier and model resolution. Each stage leaves an artifact before the next
begins.

The motion release supports rebuilding the visual records from the public motion table: the rebuild script
recomputes body and camera settings from each \texttt{clip\_id} and reruns the public renderer, without ARDY.
It rebuilds rendered frames from released motions; it does not regenerate the source motions. The verifier was
run over all 514,800 $128^2$ frames of the pre-curation generation---a superset containing every one of the
482,400 released frames: every motion and metadata label matches; color and segmentation
pixels agree on 99.88\% and 99.92\% of comparisons; joint visibility, the least stable field, agrees on
99.50\%; depth and normals exceed 99.99\%. A 360-frame check against the released $64^2$ tier passes its
declared agreement thresholds. The same script can rerender with seeded overrides to body scale, stroke, or
camera, recorded in the manifest.

Reference runs were measured on one NVIDIA RTX PRO 6000 Blackwell (96\,GB). The 30k-step image stage ran at
0.10\,s/update with a 10.5\,GB peak. The final factorized image- and random-initialized video runs each ran
at 0.64\,s/update and peaked at 37.8 and 37.7\,GB; the local-mixer run took 1.24\,s/update and peaked at
44.1\,GB. We did not test the same configurations on lower-memory GPUs. The runs used different optimizer
kernels, so these measurements document run cost rather than a controlled architecture-speed comparison
and transfer only approximately to other hardware.

Our code is MIT and the dataset is CC0-1.0. Motions were generated with ARDY's 20-fps Core model. ARDY's
code is Apache-2.0; its checkpoints are distributed under the
\href{https://www.nvidia.com/en-us/agreements/enterprise-software/nvidia-open-model-agreement/}{\color{black}NVIDIA Open
Model Agreement}, which permits derivative works and does not assert ownership of model outputs. The released
frames are procedurally rendered and the released source motions are ARDY model outputs from text prompts; no
directly captured human images or motion sequences are included in this release.

\section{Limitations and future work}\label{sec:future}
The domain is narrow by design: one raster style, one figure, orthographic cameras, no real-video proxy. The
color-based evaluator measures visible topology, not bone
length, joint angles, left/right semantics, or biomechanics; its measured blind spots include whole-limb
swaps, proportion changes, and time reversal. The arm-chain localization study covers one topology-preserving motion edit; localization for other body
parts and failure types remains untested. The
motions inherit ARDY's capabilities and biases; curation
was prompt-level over inspected realisations from every seed, and we have not human-validated every remaining
motion. The seed-disjoint test measures in-domain generation under unseen motion realisations of known
prompts, not zero-shot prompts or human motion at large, and the color metrics do not establish prompt
adherence. Each model row in Table~\ref{tab:backbone-results} is a single training run evaluated with three
sampling seeds; those seeds measure sampling variation, not training variance, and no row carries a scaling claim.

A learned rig estimator is future work that follows from these blind spots: mapping a generated video back to
a rig and re-rendering it could expose rig-space failures the color metrics miss, such as unstable bone
length or end-effector errors. Its scores would need validation on controlled malformations before use as
evidence.

\paragraph{Broader impact.}
The release lowers the compute and data-access barrier to studying video generation and makes failures easier
to inspect without collecting images of people. Its narrow synthetic domain is also a risk if results are
presented as evidence about realistic human video: the motions inherit the source model's biases, and the
color-based diagnostics are not general video-quality or human-motion scores. We therefore document their
measured blind spots, scope all claims to this dataset and protocol, and release the generator and evaluator so
that reported results can be inspected rather than treated as opaque rankings.

\section{Conclusion}
Dancing Stick Figures provides a compact environment for training and evaluating small, domain-specific video-generation models end to end.
Each rendered video is released with its source rig and generation parameters, together with the harness
needed to generate additional data. The release also includes reference models and dataset-specific
measurements of topology, part identity, color purity, and motion. These measurements are intended as
diagnostic tools: they help identify what a model has learned, where its outputs break down, and whether a
change in training or architecture improves a particular failure mode.

Together, these components support a complete, reproducible workflow for video-generation research. A
researcher can generate or modify the data, train a 64-frame video model, inspect its outputs, measure
specific failures, revise the model, and repeat the process at a manageable scale. The present dataset
deliberately focuses on text-conditioned videos of a single rigged figure, but the same workflow can be
extended with new motions, appearances, camera conditions, and diagnostics. By releasing the data,
generation harness, reference models, and evaluator together, we aim to make video-generation research
easier to enter, reproduce, and study through iteration.

\end{document}